\documentclass[sigconf]{acmart}
\AtBeginDocument{%
  }

\setcopyright{acmlicensed}
\copyrightyear{2018}
\acmYear{2018}
\acmDOI{XXXXXXX.XXXXXXX}
\acmConference[Conference acronym 'XX]{Make sure to enter the correct
  conference title from your rights confirmation email}{June 03--05,
  2018}{Woodstock, NY}
\acmISBN{978-1-4503-XXXX-X/2018/06}

\usepackage{algorithm}
\usepackage{algorithmicx}%
\usepackage{algpseudocode}%
\usepackage{amsmath}
\usepackage{amsfonts}
\usepackage{subcaption}
\usepackage{enumitem}
\usepackage[table]{xcolor}

\begin{document}

%%
%% The "title" command has an optional parameter,
%% allowing the author to define a "short title" to be used in page headers.

%%
%% The "author" command and its associated commands are used to define
%% the authors and their affiliations.
%% Of note is the shared affiliation of the first two authors, and the
%% "authornote" and "authornotemark" commands
%% used to denote shared contribution to the research.

\title{Holistic Explainable AI (H-XAI): Extending Transparency Beyond Developers in AI-Driven  Decision Making}

\author{Kausik Lakkaraju}
\affiliation{%
 \institution{AI Institute, University of South Carolina}
 \city{Columbia}
 \state{South Carolina}
 \country{USA}
 }

\author{Siva Likitha Valluru}
\affiliation{%
 \institution{AI Institute, University of South Carolina}
 \city{Columbia}
 \state{South Carolina}
 \country{USA}
 }

\author{Biplav Srivastava}
\affiliation{%
 \institution{AI Institute, University of South Carolina}
 \city{Columbia}
 \state{South Carolina}
 \country{USA}
 }

%%
%% By default, the full list of authors will be used in the page
%% headers. Often, this list is too long, and will overlap
%% other information printed in the page headers. This command allows
%% the author to define a more concise list
%% of authors' names for this purpose.
\renewcommand{\shortauthors}{Lakkaraju et al.}

%%
%% The abstract is a short summary of the work to be presented in the
%% article.
\begin{abstract}
As AI systems increasingly mediate decisions in domains such as credit scoring and financial forecasting, their lack of transparency and bias raise critical concerns for fairness and public trust. Existing explainable AI (XAI) approaches largely serve developers, focusing on model justification rather than the needs of affected users or regulators. We introduce Holistic eXplainable AI (H-XAI), a framework that integrates causality-based rating methods with post-hoc explanation techniques to support transparent, stakeholder-aligned evaluation of AI systems deployed in online decision contexts. H-XAI treats explanation as an interactive, hypothesis-driven process, allowing users, auditors, and organizations to ask questions, test hypotheses, and compare model behavior against \textit{automatically} generated random and biased baselines. By combining global and instance-level explanations, H-XAI helps communicate model bias and instability that shape everyday digital decisions. Through case studies in credit risk assessment and stock price prediction, we show how H-XAI extends explainability beyond developers toward responsible and inclusive AI practices that strengthen accountability in sociotechnical systems.
\end{abstract}

\maketitle

% -------------
\section{Introduction}
\begin{figure}[h]
    \centering
    \begin{subfigure}[b]{0.2\textwidth}
        \centering
        \includegraphics[width=\textwidth]{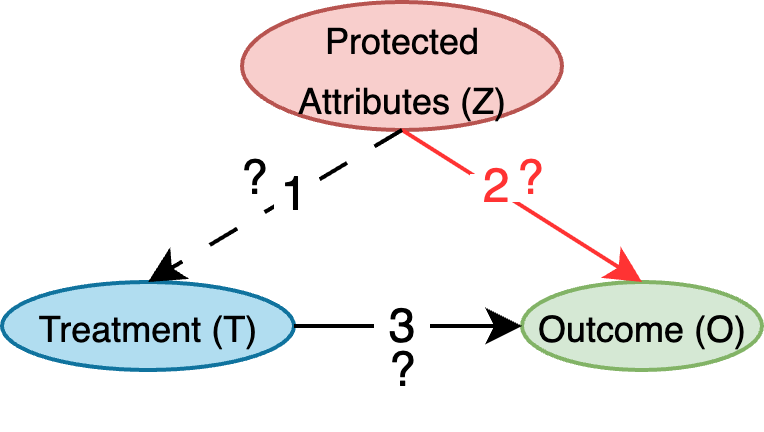}
        \vspace{-1.5em}
        \caption{Causal Graph}
        \label{fig:gen-cm}
    \end{subfigure}
    \begin{subfigure}[b]{0.25\textwidth}
        \centering
        \includegraphics[width=\textwidth,height=2cm]{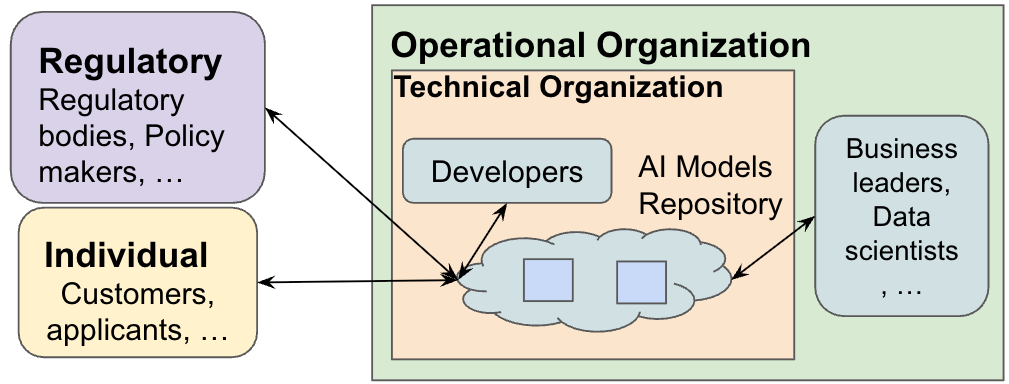}
        \caption{AI Stakeholders}
        \label{fig:stakeholders}
    \end{subfigure}
    \hfill
    \begin{subfigure}[b]{0.5\textwidth}
        \centering
        \includegraphics[width=\textwidth,height=5cm]{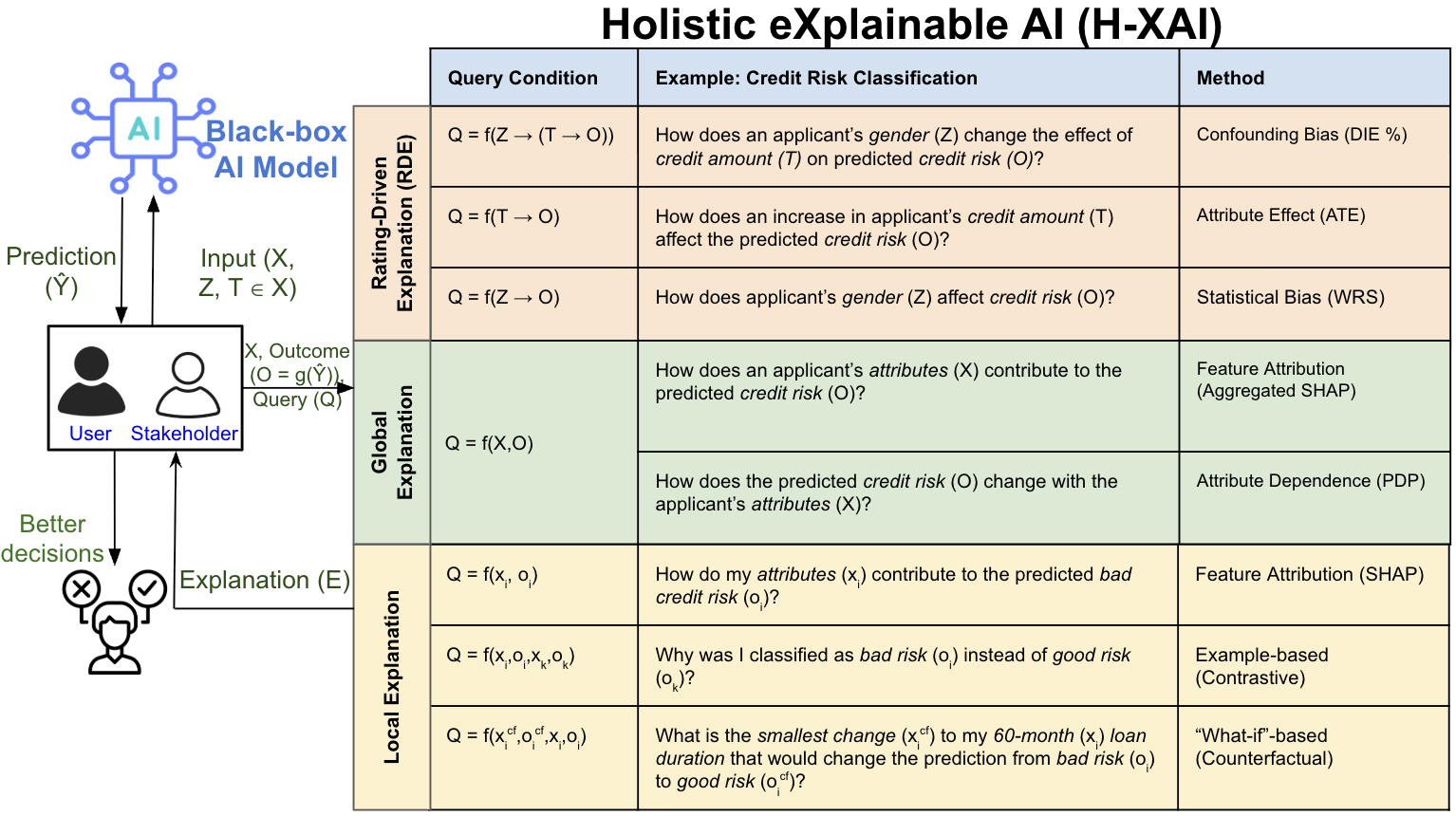}
        \caption{H-XAI Workflow}
        \label{fig:xai-workflow}
    \end{subfigure}
    \caption{(a) Generalized Causal Graph; (b) Stakeholder types in the context of Blackbox AI; (c) \textit{H-XAI} Workflow: End-user receives output from a blackbox AI. Stakeholders may seek related explanations. ``Method'' shows type (.) and instances from case studies.}
    \label{fig:stakeholder-xai}
\end{figure}

Explainable AI (XAI) has long promised to help people understand how AI systems behave. Yet despite the variety of methods available, most existing tools are still designed for developers or model builders, not for the wider range of stakeholders who interact with or are affected by AI systems \cite{mueller2019explanation,bhatt2020machine,deshpande2022responsible}. Many such systems now appear in settings where decisions are delivered through online interfaces or web-based services,
which further broadens who may need to interpret a system's behavior. Figure \ref{fig:stakeholders} illustrates three key stakeholder groups: individual users (e.g., applicants using online decision tools), operational organizations (e.g., data science teams), and regulatory bodies.

Consider the task of credit risk assessment. An applicant might ask, ``Would my loan have been approved if I had requested a smaller amount?'' A data scientist may wonder, ``Which applicant features drive approval decisions on average?'' A regulator might want to know whether a model systematically favors certain demographic groups. These questions differ in scope and cannot be handled by a single explanation method. Current XAI tools typically focus on justifying a model's output rather than helping stakeholders examine it from different angles. They rarely support moving between alternative hypotheses \cite{hoffman2023explainable}, testing counterfactuals \cite{miller2023explainable}, or identifying when a model behaves unreliably \cite{suresh2021beyond}. They also tend to produce a fixed explanation, even though practical use often requires several rounds of inspection \cite{hoffman2023increasing}.

Past work on black-box model rating offers one way to study model behavior without access to training data. \cite{srivastava2018towards} introduced a two-step procedure to quantify bias in AI services, followed by visualization tools for exploring accuracy-fairness trade-offs \cite{vega-tool}. These ideas have since been applied to chatbots \cite{srivastava2020personalized} and search engines \cite{tian2023mitigating}. While such ratings communicate how a system behaves ``in the wild,'' they do not answer the questions stakeholders often ask about why a particular decision changed or how sensitive the model is to certain inputs.

Causal fairness methods go a step further by isolating the influence of protected attributes on predictions \cite{carey2022causal}. This mirrors the kinds of ``what if this were different?'' questions people tend to ask. Building on this line of work, \cite{kausik2022why,srivastava2023advances} proposed a causal rating framework applied across domains ranging from sentiment analysis \cite{kausik2024rating} to time-series forecasting \cite{lakkaraju2024timeseries,lakkaraju2025creating}. These methods quantify statistical and confounding bias and evaluate model stability under perturbations. However, they do not provide the instance-level explanations often needed in
operational contexts.

In this paper, we propose a framework, \textit{Holistic XAI (H-XAI)}, that brings together traditional XAI techniques and \textit{rating-driven explanations} (RDE). RDE is helpful for questions involving model instability, bias, and comparisons across models; traditional methods such as SHAP, PDPs, and counterfactuals address local explanations and feature-level reasoning. H-XAI lets stakeholders move between these viewpoints depending on the question at hand. To illustrate how this works, we present two case studies covering six concrete scenarios: credit risk classification using
the German Credit dataset and financial time-series forecasting with stock prices. Across these scenarios, we show how the combination of methods supports the kinds of questions raised by applicants, regulators, and data scientists.

Our contributions are as follows:
\begin{enumerate}
    \item We introduce \textit{Holistic XAI (H-XAI)}, a formulation combining black-box ratings with post-hoc XAI techniques for both instance- and system-level analysis (Figure \ref{fig:xai-workflow});
    \item We develop a \textit{rating-driven explanation} (RDE) workflow that supports hypothesis testing with causal metrics and comparative evaluation using automatically generated random and biased baselines (Figure \ref{fig:rde-workflow});
    \item We demonstrate how H-XAI addresses different stakeholder questions across binary classification and time-series forecasting (Section \ref{sec:cases}), and discuss practical method-selection strategies (Table \ref{tab:method-map}).
\end{enumerate}

\section{Background and Related Work}
In this section, we describe how we categorized stakeholders by drawing on prior work, and how we selected a subset of these stakeholder types to represent in our case studies in Section \ref{sec:cases}. We also explain how the questions we chose reflect the explanation needs identified in the literature, and conclude by situating our work within relevant prior work. 
% An extended discussion of related work, including a dedicated section on explainability for time-series models, is available in the appendix (Section \ref{app:related-work}).

\subsection{Background: Stakeholders and Their Explanation Needs}
\label{sec:bgd}

We follow the stakeholder taxonomy from \cite{deshpande2022responsible}, which identifies three main groups (Figure \ref{fig:stakeholders}):

\noindent \textbf{1. Individual stakeholders}: End users directly impacted by model predictions. Their questions often focus on personal outcomes and decision justification.
    
\noindent \textbf{2. Regulatory stakeholders}: Policy-makers and auditors concerned with fairness, bias, and legal compliance.
    
\noindent \textbf{3. Organizational stakeholders}: Includes \textbf{operational roles} (e.g., business leads, data scientists) and \textbf{technical roles} (e.g., developers) concerned with performance and reliability.

To structure their explanation needs, we adapt the XAI question taxonomy from \cite{liao2021question}, which organizes nine categories of explanatory questions from interviews with industry practitioners. Though originally domain-agnostic, many of their core questions, such as ``Why was this prediction made?'', ``What would change the outcome?'', or ``Which features matter most?'', align closely with our stakeholder scenarios in credit scoring and forecasting. We pair each question type with a fitting method: SHAP for feature attribution, counterfactuals for what-if queries, and RDE for behavior under perturbations or across subgroups. A full mapping is provided in the appendix (Table \ref{tab:question_mapping}).
%A detailed mapping of our scenario questions to the categories of the XAI Question Bank is provided in the appendix (Table \ref{tab:question_mapping}).

\subsection{Explainability in AI}

Explainable AI (XAI) methods aim to clarify models behavior by describing predictions, decision patterns, and potential risks \cite{arrieta2020explainable}. Toolkits like AIX360 \cite{arya2019one} offer diverse techniques to support this goal. However, most explanations remain static, built to justify outcomes rather than support interaction or hypothesis testing. Prior work argues that explanation is not a fixed output but a process of exploration \cite{mueller2019explanation, hoffman2023increasing}. Users often need to ask follow-up questions or test alternate scenarios to build meaningful understanding \cite{abdul2018trends, chromik2021human}. Existing methods rarely support this kind of iterative reasoning or sensitivity analysis. In \textit{H-XAI}, we use post-hoc methods like SHAP and PDPs to answer questions about feature influence on predictions, both locally and globally. They, however, are limited when users want to test hypotheses, such as whether a model's decision would change under a counterfactual scenario, or whether outcomes differ systematically across groups. For such cases, we use RDE that estimate perturbation effects, account for confounding, and provide model comparisons.

\subsection{Rating of AI Models}

Prior work has introduced rating methods to assess AI models from a third-party perspective, often without access to training data. \cite{srivastava2020rating} proposed a framework to quantify bias, particularly gender bias, in machine translation systems \cite{srivastava2018towards}, using visualizations to communicate results \cite{vega-tool,vega-userstudy-translatorbias}. However, these approaches lacked a causal foundation. Recent work introduced causal rating methods to isolate the effects of protected attributes on model outcomes. This was first applied to sentiment analysis systems \cite{kausik2024rating} and then extended to composite NLP tasks \cite{kausik2023the} and time-series forecasting models \cite{lakkaraju2024timeseries,lakkaraju2025creating}. These methods go beyond correlation by estimating treatment effects and adjusting for confounding. In this paper, we build on that foundation but shift the role of ratings: rather than treating them solely as evaluation metrics, we use them to explain model behavior. By integrating ratings with traditional XAI methods, we allow stakeholders to compare models, test hypotheses, and interpret robustness and bias in a more structured way.

\subsection{Understanding Stakeholder Needs in XAI}
Understanding stakeholder needs is critical for building explanations that are useful, not just technically accurate \cite{bhatt2020machine}. For example, \cite{bundas2023facilitating} highlights NIST's emphasis on hierarchical vocabularies: high-level concepts should be broadly accessible, while lower levels serve expert users. Studies show stakeholders prefer interactive explanations. Interviews in \cite{hoffman2023explainable} show a strong preference for tools that allow input manipulation and scenario exploration. RDE supports this by allowing users to perturb inputs and observe  causal effects on model behavior (Figure~\ref{fig:ts-e3}). While current XAI methods focus on local explanations, users often want a broader understanding of model behavior \cite{mueller2019explanation}. \cite{suresh2021beyond} proposes a three-level framework for interpretability needs, including the need to understand model limitations. RDE addresses this by showing how behavior changes under different conditions, supporting users with a structured robustness evaluation.
In this paper, we show how \textit{H-XAI} brings these ideas together, supporting interactive, stakeholder-aligned explanations through a combination of rating and post-hoc XAI methods.

\section{Approach}
\label{sec:approach}

\begin{figure}[h]
    \centering
    \includegraphics[width=0.45\textwidth, height=3cm]{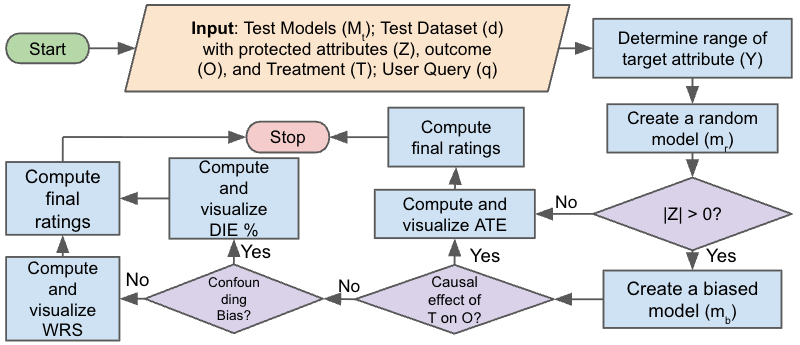}
    \caption{RDE Workflow.}
    \label{fig:rde-workflow}
\end{figure}

To explain model behavior in ways that support robustness, fairness, and practical decision-making, we propose a hybrid framework called \textit{Holistic-XAI (H-XAI)}. It integrates traditional post-hoc explanation methods with rating-driven explanation (RDE) approach. Unlike approaches that rely solely on post-hoc explanations, \textit{H-XAI} combines both: traditional XAI methods are well-suited for explaining individual predictions and feature attributions, while RDE supports broader questions about robustness, bias, and perturbation effects. This combination lets stakeholders examine both how individual predictions are made and how the model behaves under different input conditions.

\subsection{Rating-Driven Explanation (RDE)}
\label{sec:rde}

RDE combines three components: (a) a causal model that specifies the structural assumptions between key attributes; (b) a set of metrics to quantify causal and statistical bias; and (c) a workflow that connects user queries to ratings and model comparisons.

\subsubsection{Causal Model}

We begin by specifying a causal model for the task, which includes assumptions about how attributes influence one another. Figure \ref{fig:gen-cm} shows the generalized causal diagram used across our case studies. While the causal model captures the underlying assumptions, the diagram is its visual representation, making explicit the directional relationships we test. In this setup, the outcome \( O \) (e.g., prediction or residual error) is influenced by a treatment variable \( T \) (e.g., credit amount, input perturbation), which may itself be influenced by a protected attribute \( Z \) (e.g., gender and age). If \( Z \) is a common cause of both \( T \) and \( O \), then confounding exists and must be accounted for to estimate a valid treatment effect.

This diagram allows us to define target causal estimands such as \( E[O \mid do(T)] \) (i.e., the deconfounded distribution after adjustment), which represents the expected model outcome under an intervention on \( T \). In practice, causal models are often specified by experts, derived from controlled experiments, or inferred using causal discovery algorithms \cite{gong2024causal,varando2024pairwise}, which themselves rely on statistical associations in the observed data. We begin with a hypothesized causal model that assumes protected attributes act as confounders and explore different scenarios within this structure. Even when based on assumptions, such models provide a structured framework for reasoning under interventions, something that correlational methods often fail to do, especially in the presence of confounding (as illustrated by Simpson's paradox \cite{simpson2022judea}). For instance, in Section \ref{sec:case-study-1}, we consider the possibility that a loan applicant's status influences the credit amount requested, an entirely plausible scenario in practice. The outcome $O$ may be either the model prediction $\hat{Y}$ or a derived quantity $g(\hat{Y})$. For example, in time-series settings, we define $O = |\hat{Y} - Y|$ as the residual error. This lets us ask: ``Does a certain input perturbation systematically worsen performance for a particular group?''.

\subsubsection{Rating metrics}
\label{sec:rating-metrics}

While RDE builds on prior rating methods, it repositions their function. Earlier works primarily used ratings to compare model performance across input conditions. RDE instead uses them as part of a structured explanation workflow: formulating a question (e.g., ``Does treatment \textit{T} affect outcome \textit{O}?''), selecting a baseline, applying a metric, and interpreting the result as a potential explanation. This gives ratings an explanatory role, especially in helping stakeholders reason about model bias, instability, or robustness under different conditions. We use three metrics, each designed to answer a specific kind of question. The full implementation details are in Appendix \ref{app:rde-metrics}.

\medskip
\noindent \textbf{Weighted Rejection Score (WRS).}
Let $Z$ have $m$ groups and let $\mathcal{C}=\{c_1,\dots,c_K\}$ be a set of confidence levels (e.g., 95\%,75\%,60\%), with corresponding weights $w_k$. For each unordered pair $(a,b)$ of groups and each confidence level $c_k$ compute the two-sample $t$-test statistic $t_{ab}^{(k)}$ and define the indicator
\[
I_{ab}^{(k)} = 
\begin{cases}
1 & \text{if } |t_{ab}^{(k)}| > t_{\mathrm{crit}}(c_k, \,\nu_{ab}^{(k)}) \\
0 & \text{otherwise,}
\end{cases}
\]
where $t_{\mathrm{crit}}(c_k,\nu)$ is the critical $t$-value at confidence $c_k$ with degrees of freedom $\nu_{ab}^{(k)}$. The WRS is the weighted sum of these rejections:
\[
\mathrm{WRS} \;=\; \sum_{k=1}^K w_k \sum_{1\le a < b \le m} I_{ab}^{(k)} .
\]
In our experiments we use $\mathcal{C}=\{0.95,0.75,0.60\}$ with weights $w=\{1.0,0.8,0.6\}$; the inner sum runs over all $\binom{m}{2}$ group pairs.

\medskip
\noindent \textbf{Average Treatment Effect (ATE).}
When the question is about the effect of a specific change (e.g., halving the requested loan amount), we estimate the ATE,
\[
\text{ATE} \;=\; \mathbb{E}[O \mid \text{do}(T=1)] \;-\; \mathbb{E}[O \mid \text{do}(T=0)],
\]
which is the average change in the model outcome under the intervention. In practice we estimate this with propensity-score matching (PSM) or G-computation when the treatment is continuous. Small ATE means the model is effectively insensitive to that change; a large ATE means the change shifts predictions or errors.

\medskip
\noindent \textbf{Deconfounded Impact Estimation (DIE\%).}
An observed ATE can be misleading if a protected attribute $Z$ affects both $T$ and $O$. To quantify that distortion we compute
\[
\text{DIE\%} \;=\; 100 \times \frac{\lvert \text{ATE}_{\text{unadj}} - \text{ATE}_{\text{adj}}\rvert}{\lvert \text{ATE}_{\text{adj}}\rvert},
\]
where $\text{ATE}_{\text{adj}}$ is the ATE after adjusting for $Z$ (via PSM or G-computation) and $\text{ATE}_{\text{unadj}}$ is the ATE before adjusting. If DIE\% is large, much of the apparent effect of $T$ is actually explained by group differences; if DIE\% is small, the treatment effect is robust to confounding.

\medskip
These three measures play distinct roles in explanation: WRS measures distributional disparities across groups; ATE quantifies how much an intervention changes model outputs on average; DIE\% quantifies the effect of confounders on the relationship between treatment and the outcome. We selected these metrics because they are straightforward to explain to non-technical stakeholders (regulators, applicants) and because together they let us rapidly triage whether a model's behavior warrants a deeper, instance-level analysis (for example, using SHAP or counterfactuals).

\subsubsection{RDE Workflow}

Figure \ref{fig:rde-workflow} shows the end-to-end RDE process. Given a user query (e.g., ``Is model performance consistent across demographic groups?''), we:
1. Define a treatment ($T$), outcome ($O$), and protected attribute ($Z$),
2. Select the appropriate metric (e.g., ATE or DIE \%),
3. Compute scores for the target / test model,
4. Compare it to two automatically generated reference models:
   - A \textbf{random model}, which outputs predictions at random,
   - A \textbf{biased model}, which gives different, often extreme predictions based solely on the value of the protected attribute $Z$. For example, in the context of credit risk assessment, always classifying one group as good risk and another group as bad risk.

This comparison grounds the explanation. A model that behaves closer to the biased baseline may raise fairness concerns; one that matches the random model may be unreliable. These comparisons are not just for evaluation: they help users understand \textit{why} the model's behavior is problematic and where. RDE outputs can be numeric scores or visual plots, depending on the user query. For example, Figure \ref{fig:gc-e2} shows a case where RDE highlights age-based confounding in a credit classifier, while Figure \ref{fig:ts-e3} shows how error rates change under missing input perturbations.

\subsection{Traditional XAI Methods Used}
\label{sec:xai}

While RDE provides explanations concerning the robustness of the models at a global level, it is not suitable for local feature attribution or communicating the robustness of the model at a global level. To address these needs, we integrate standard post-hoc explanation methods, which stakeholders may already be familiar with. 

% The complete definitions of each can be found in Appendix \ref{app:xai-methods}.

\noindent \textbf{Partial Dependency Plots (PDPs):} Show the marginal effect of a feature on predictions by averaging over the dataset. Useful for understanding monotonicity or directionality, but unreliable when features are correlated. In Figure \ref{fig:gc-e2}, we use PDPs to examine the effect of credit amount across applicant groups.

\noindent \textbf{SHapley Additive exPlanations (SHAP):} Estimates the marginal contribution of each feature to a prediction, using additive Shapley values. For time-series models, we use TreeSHAP on surrogate models \cite{raykar2023tsshap} to reduce computational cost. This makes the explanations more scalable, even for long input sequences. In our case study, we evaluated the surrogate on a small subset (10\% of data, 60 samples) and found that its MASE and SMAPE differed from MOMENT's by only 0.66 and 0.18\%, respectively (see Table \ref{tab:mase-smape-surrogate} in the appendix).

\noindent \textbf{Counterfactuals:} Identify minimal input changes needed to flip a prediction \cite{wachter2017counterfactual}. This is useful for questions like, ``What should this applicant change to get approved?''

\subsection{Holistic eXplainable AI (H-XAI)}

\textit{H-XAI} brings together traditional post-hoc methods, some of which explain individual predictions (like SHAP and counterfactuals) and others that show general patterns (like PDPs or global SHAP), with RDE, which focuses on robustness by testing how model behavior changes under interventions and comparing it to biased and random baselines. This lets us analyze not just how the model works, but how stable and fair it is under different conditions. This process is often driven by stakeholders themselves, who have hypotheses involving certain variables they want to test. Stakeholders may begin with a general question (``Is the model fair across groups?''), use RDE to detect problems, then drill down into specific examples using SHAP or counterfactuals to understand the source of those issues. As shown in Figure \ref{fig:xai-workflow}, this setup allows users to move between different types of explanations depending on the question, without being locked into a single method or interpretability style. We demonstrate this combined framework through two use cases in Section \ref{sec:cases}, mapping specific stakeholder questions to the methods used to answer them.

\begin{figure}
     \centering
     \begin{subfigure}[b]{0.20\textwidth}
         \centering
    \includegraphics[width=\textwidth]{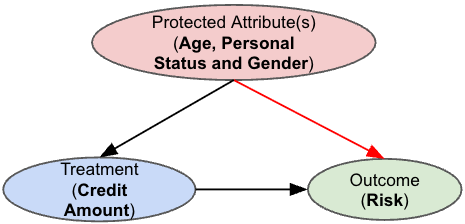}
         \caption{~} 
         \label{fig:gc-cm}
     \end{subfigure}
     % \hfill
     \begin{subfigure}[b]{0.24\textwidth}
     \centering
     \includegraphics[width=\textwidth]{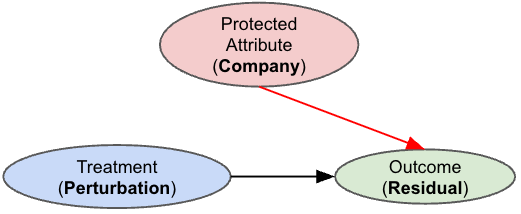}
     \caption{~}
     \label{fig:ts-cm}
     \end{subfigure}
    \caption{(a) Causal diagram for binary classification on the German Credit dataset. Protected attributes include \textit{Age}, \textit{Personal Status, Gender}, or their combinations, depending on query.
    (b) Causal diagram for time-series forecasting (stock prices), adapted from \cite{lakkaraju2024timeseries}.}
    \label{fig:cm}
    \vspace{-1em}
\end{figure}
\section{Case Studies}
\label{sec:cases}

\begin{figure}
    \centering
    \includegraphics[width=0.40\textwidth, height=4cm]{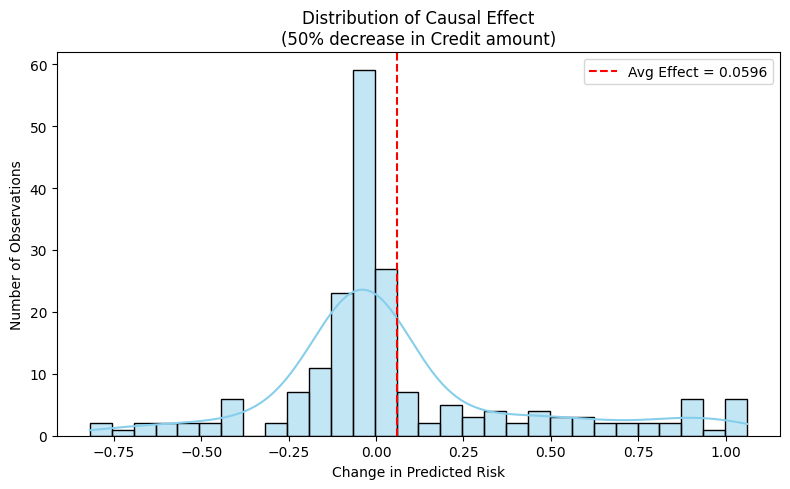}
    \caption{\textbf{Hypothesis:} Does decreasing the credit amount by 50\% improve the chances of loan approval? 
    \textbf{Conclusion:} On average, decreasing the requested credit amount by 50\% increases the risk by 0.06, meaning applicants will be more likely to be classified as \textit{good risk} (denoted by class 1). This suggests that asking for a smaller loan may improve the likelihood of loan approval only marginally.}
    \label{fig:gc-1}
\end{figure}
\vspace{-0.5em}
\begin{figure}
    \centering
    \includegraphics[width=0.40\textwidth, height=3.5cm]{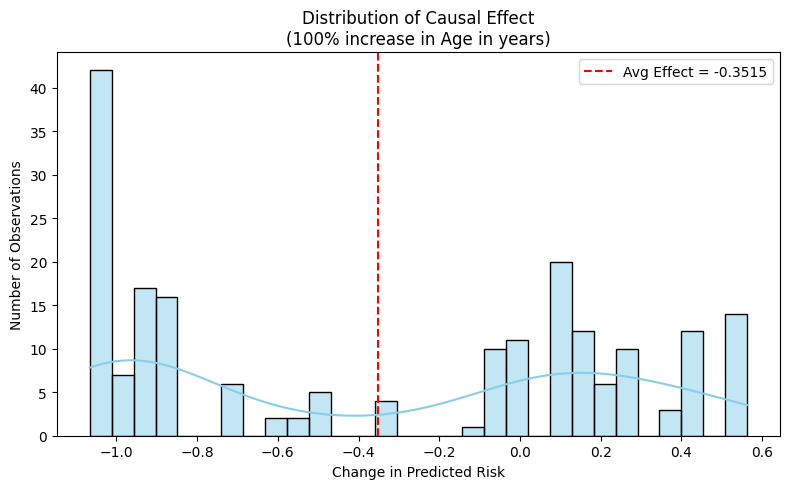}
    \caption{\textbf{Hypothesis:} Does doubling applicants' age, reduce their chances of being approved for a loan?
    \textbf{Conclusion:} On average, doubling a person’s age led to a noticeable drop in the predicted creditworthiness (risk score decreased by 0.35). This suggests that the model tends to classify older applicants as higher risk, which could be a sign of age-related bias.}
    \label{fig:gc-2}
\end{figure}

\begin{figure}
    \centering
    \includegraphics[width=0.40\textwidth, height=3.5cm]{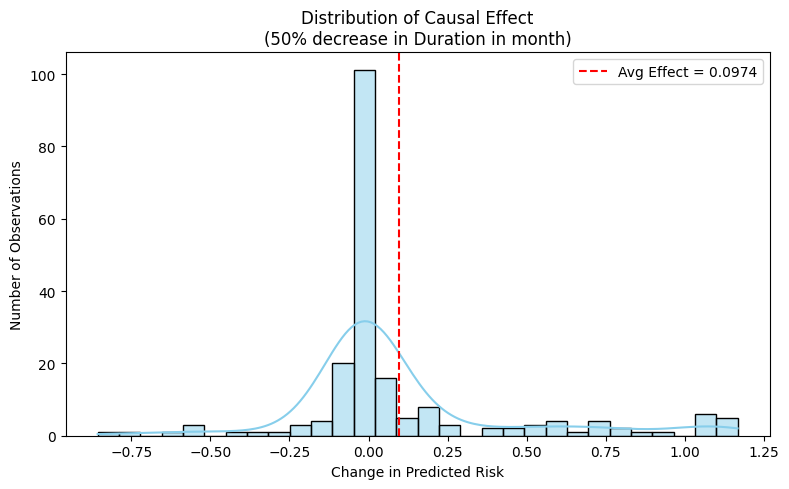}
    \caption{\textbf{Hypothesis:} Does decreasing the loan duration by half improve the chances of loan approval?
    \textbf{Conclusion:} Reducing the loan duration by half led to a slight increase in predicted creditworthiness (risk score increased by 0.097). This indicates that the model tends to view shorter loan durations more favorably, which may slightly help an applicant get approved.}
    \label{fig:gc-3}
\end{figure}

\begin{figure*}
    \centering
    \includegraphics[width=0.85\textwidth, height=7.5cm]{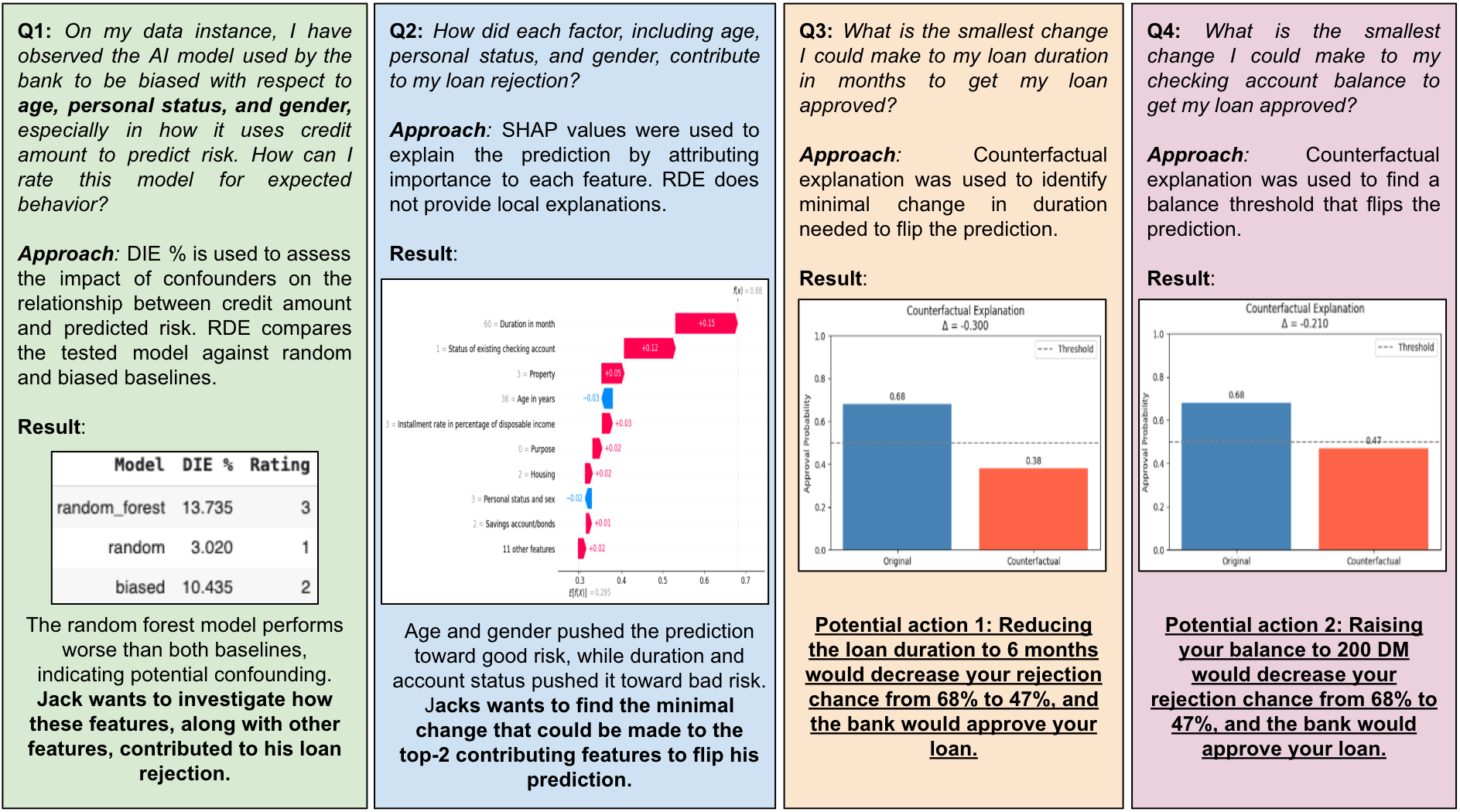}
    \caption{\textbf{Stakeholder}: Jack (applicant; individual); \textbf{Scenario-1}: Bank XYZ uses a \textbf{Random Forest} model to classify applicants as \textit{good} or \textit{bad risk} based on features like credit amount, age, personal status, gender, and more. Jack was \textbf{classified as a \textit{bad risk}} and wants to understand why, and what changes he could make to be considered a good risk and get his loan approved.}
    \label{fig:gc-e1}
\end{figure*}
\begin{figure}
    \centering
    \includegraphics[width=0.47\textwidth]{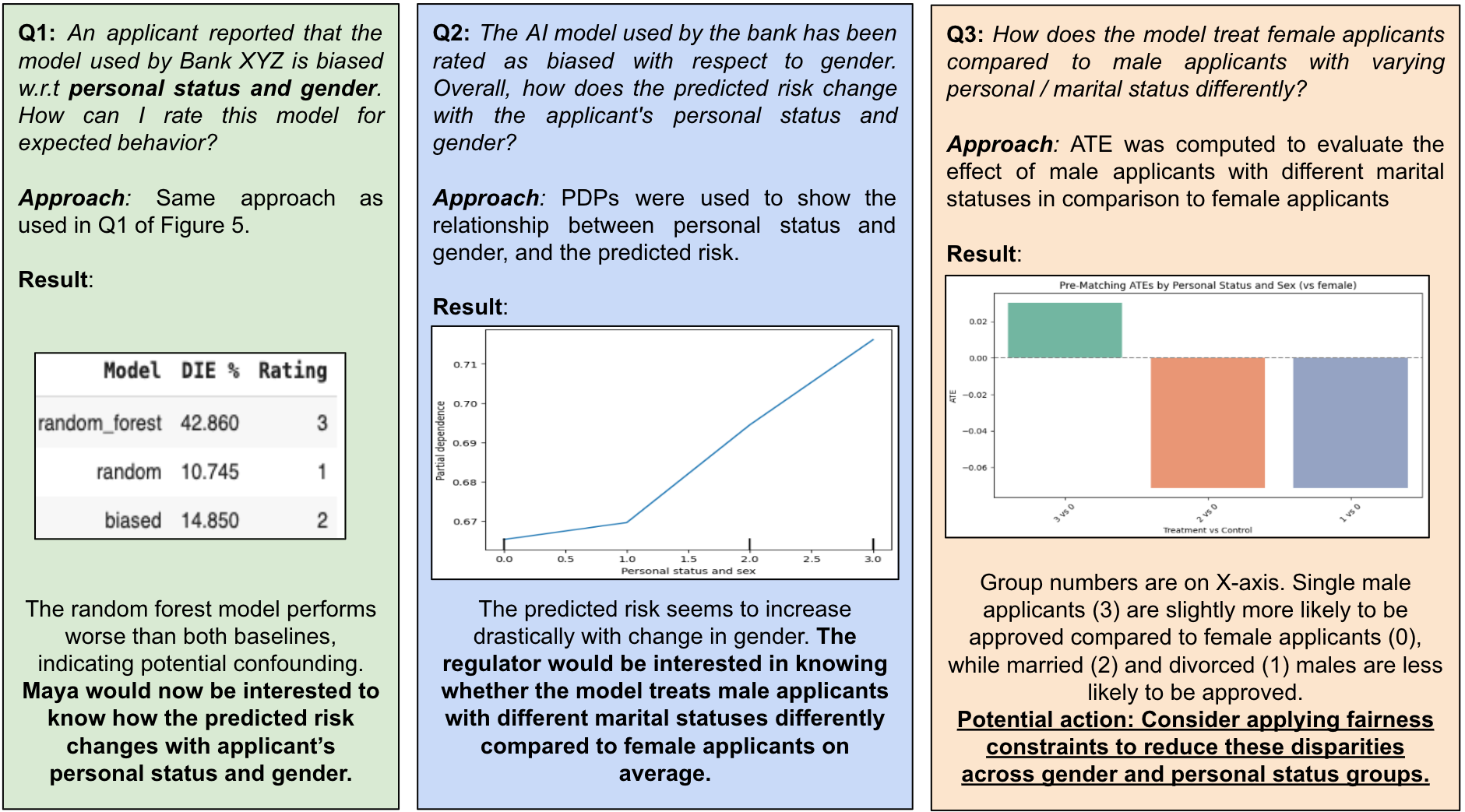}
    \caption{\textbf{Stakeholder:} Maya (Regulatory official; Regulatory body). \textbf{Scenario-2:} An applicant reports that the AI model used by Bank XYZ is biased with respect to personal status and gender, and suspects their loan was rejected for this reason. Maya, a regulator from the ABC Bureau, begins an investigation to determine whether this bias is present.}
    \label{fig:gc-e2}
\end{figure}
\begin{figure}
    \centering
    \includegraphics[width=0.47\textwidth]{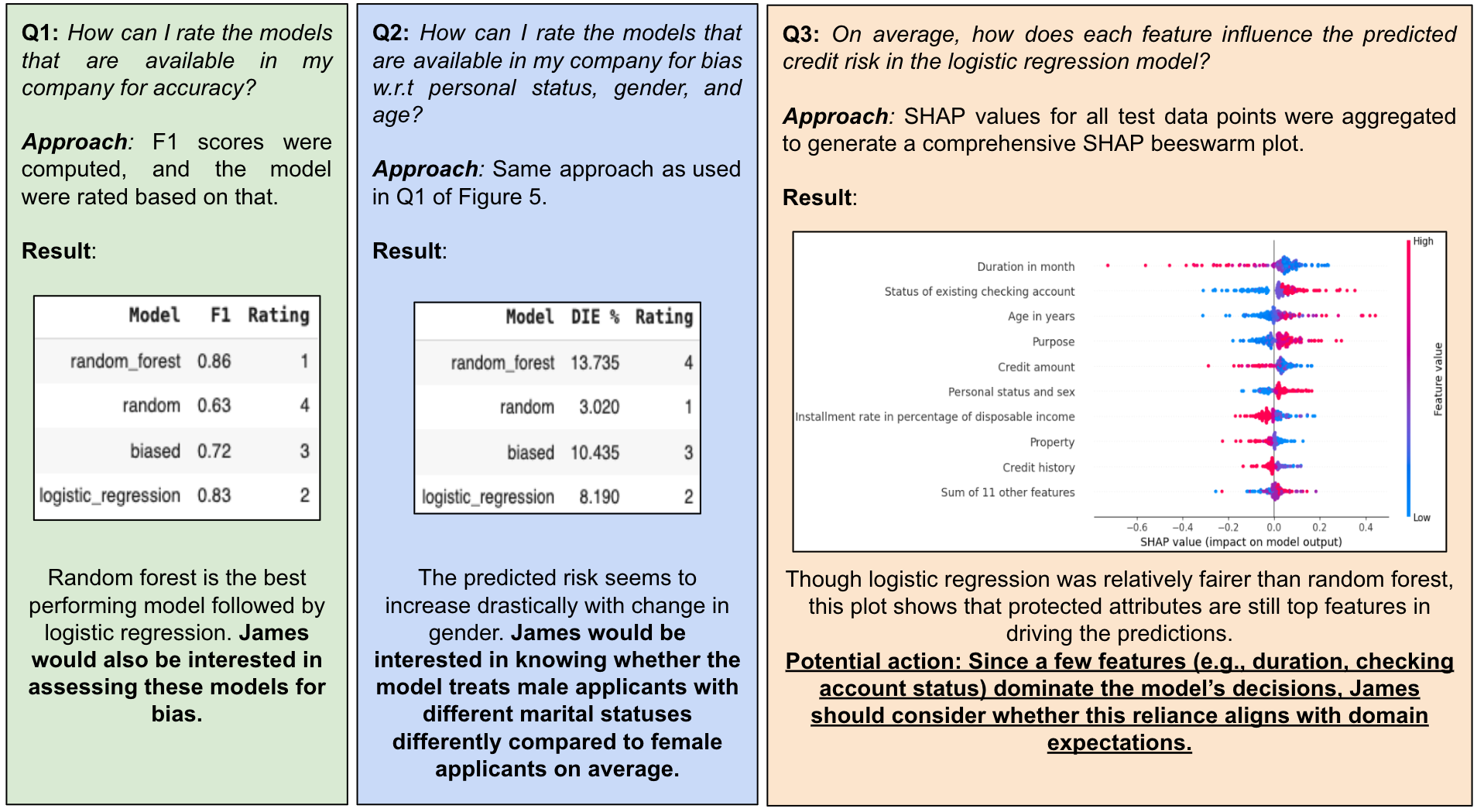}
    \caption{\textbf{Stakeholder:} James (Data Scientist; Operational Organization). \textbf{Scenario-3:} Given a set of candidate models, James sets out to evaluate which one performs best on the client’s credit risk data. He plans to assess both accuracy and fairness before selecting a model for deployment; Interestingly, the global SHAP summary (Q2) is closely aligned with the Jack's SHAP values shown in Figure \ref{fig:gc-e1}!}
    \label{fig:gc-e3}
\end{figure}

\begin{table*}[h!]
{\tiny
\begin{tabular}{|p{4.5cm}|p{3.2cm}|p{4.2cm}|p{4.2cm}|}
\hline
\textbf{Question} & \textbf{Method Used} & \textbf{Limitation} & \textbf{When to Combine} \\
\hline
Do outcomes differ across groups (e.g., gender, age)? & WRS & Cannot isolate cause of disparity (only statistical difference). & Combine with DIE \% to check if protected attribute has a confounding or direct effect. \\
\hline
Does a certain input (e.g., credit amount) causally influence the outcome? & ATE & Does not explain at the individual level; ignores confounding if unadjusted & Use DIE\% to validate causal effect is not confounded \\
\hline
Is the effect of a feature (e.g., credit amount) on the outcome distorted because it is influenced by a protected attribute (or confounding) (e.g., age)?  & DIE\% & Cannot be used when treatment is uniform; needs good causal estimation models; not instance-level & Use SHAP to identify where and how confounding influences predictions at the instance level \\
\hline
How does a feature influence predictions overall? & PDP & Fails when features are correlated; only shows average trends & Use ATE for more reliable estimation if confounding is present. \\
\hline
Which features contribute most to a single prediction? & SHAP (local) & Can be unstable with feature interactions; does not show global behavior & Use global SHAP or ATE to see if local pattern is consistent across dataset \\
\hline
Which features matter globally? & SHAP (global) & Attribution unreliable under correlated inputs; may mis-calculate importance & Use RDE to test if feature importance remains stable across subgroups or under input perturbations. \\
\hline
What minimal change would flip a decision? & Counterfactual Explanation & May suggest unrealistic or invalid inputs; ignores global trends & Validate counterfactual's plausibility using SHAP or PDP (helps identify features that matter to the model). \\
\hline
Does the model behave differently for one group under noisy/missing input? & DIE\% & Needs a causal model. & Use SHAP to identify which feature contributions shift across groups post-perturbation. \\
\hline
Is the model sensitive to a certain input change? & ATE (perturbation as treatment) & Only provides average effect; hides heterogeneity. & Use SHAP to see variability of effect across instances. \\
\hline
Does this error pattern resemble biased or random behavior? & RDE with biased/random baselines & Baselines are synthetic; may not match real misuse cases. & Combine with counterfactuals or SHAP to examine edge cases or deviations. \\
\hline
\end{tabular}
}
\caption{Mapping explanation methods to stakeholder questions, limitations, and when to combine them. Combinations show scenarios from our case studies (Section \ref{sec:cases}). Others are possible depending on task and data context.}
\label{tab:method-map}
\end{table*}

In this section, we present two case studies to demonstrate how \textit{H-XAI} can address the explanation needs of different stakeholders (introduced in Section \ref{sec:bgd}). These scenarios are structured around a taxonomy of questions that align with the XAI Question Bank \cite{liao2021question}. The mapping of our scenario questions to the categories of the XAI Question Bank is provided in the appendix (Table \ref{tab:question_mapping}). We show six scenarios: three for each stakeholder type in a given task.

\subsection{Case Study-1: Binary Classification}
\label{sec:case-study-1}
In the first case study, we choose binary credit risk classification as it is a well-studied, critical domain where fairness, explanation, and model behavior have real-world consequences \cite{demajo2020explainable,klosok2020towards,melsom2022explainable}. 

% \subsubsection{A. Data.}
\noindent \textbf{A. Data.}
German Credit Dataset \cite{frank2010uci} has been widely used in prior work on algorithmic fairness and explainability \cite{zemel2013learning,martins2023explainable,verma2018fairness}, making it suitable for examining how different stakeholder questions surface in a binary classification task. It contains 1,000 rows with 20 applicant features, including demographic information (e.g., \textit{age}, \textit{personal status and sex}), financial details (e.g., \textit{credit amount}, \textit{duration}, \textit{checking account status}), and other applicant-related attributes which are used to assess loan risk. We selectively analyze subsets of these features depending on the specific explanation or causal question being addressed. The primary outcome variable is binary: \( Risk \in \{0, 1\} \), where \( Risk = 1 \) indicates a good credit risk (leading to loan approval), and \( Risk = 0 \) indicates a bad credit risk (loan rejection).

% \textbf{Preprocessing:} Continuous variables such as age and credit amount were discretized into quantile-based bins for treatment effect estimation. Logistic regression and random forest, were trained to predict \( Y \) along with a random system (that assigns random predictions).

%\subsubsection{B. Causal Setup.} 
\noindent \textbf{B. Causal Setup.}
For generating RDE, we use the causal diagram shown in Figure \ref{fig:gc-cm}. Specifically, we focus on a smaller set of variables: we treat \textit{Credit Amount} as the treatment, \textit{Age} and \textit{Personal Status and Sex} as protected attributes following \cite{zemel2013learning}, and \textit{Risk} as the outcome.  In contrast, when applying XAI methods like SHAP, we do use the full feature set, since those methods are not bound by the same identifiability constraints and are designed to give feature attribution-based explanation rather than estimate causal effects.

% \subsubsection{C. AI Models Considered.} 
\noindent \textbf{C. AI Models Considered.}
We use two models: a linear, interpretable model (\textbf{Logistic Regression}), and a non-linear ensemble model (\textbf{Random Forest}). They are two of the most popular methods used for credit scoring \cite{addy2024ai}. Figure \ref{fig:gc-e3} highlights an accuracy versus confounding bias trade-off in Scenario-3. While Random Forest outperforms Logistic Regression in accuracy, it also exhibits higher confounding bias. This shows the common, but not universal, trade-off between accuracy and interpretability \cite{atrey2025demystifying}. 

We present three scenarios in which \textit{H-XAI} (described in Section \ref{sec:approach}) is applied to address questions posed by different types of stakeholders: an individual applicant (Figure \ref{fig:gc-e1}), a regulatory official (Figure \ref{fig:gc-e2}), and a data scientist within an operational organization (Figure \ref{fig:gc-e3}).

We also provide an example of a \textbf{hypothesis-driven explanation} for credit risk classification in Figures \ref{fig:gc-1}, \ref{fig:gc-2}, \ref{fig:gc-3}. We define hypothesis-driven explanations as a structured process in which users test specific assumptions or causal questions about model behavior, rather than passively receiving fixed outputs. In our formulation, such hypotheses are operationalized through causal models, allowing users to quantify effects and systematically examine how the model behaves. While this is not exactly what Miller envisioned, it aligns with the direction of his \textit{Evaluative AI} framework \cite{miller2023explainable}. The causal setup described for RDE in Section \ref{sec:rde} also supports \textbf{hypothesis-driven explanations}, allowing users to test claims about model behavior rather than receive fixed outputs, as illustrated in Figure \ref{fig:gc-1}. We show how doubling age (Figure \ref{fig:gc-2}) or halving loan duration (Figure \ref{fig:gc-3}) affects predicted \textit{credit risk}, demonstrating how hypothesis-driven explanations can serve as a standalone method for validating assumptions about model behavior. 
% \textit{Age}, \textit{Personal status}, and \textit{Gender} are assumed to act as confounders for both \textit{credit amount} and loan \textit{duration in months}.

\begin{figure}
    \centering
    \includegraphics[width=0.48\textwidth, height=5cm]{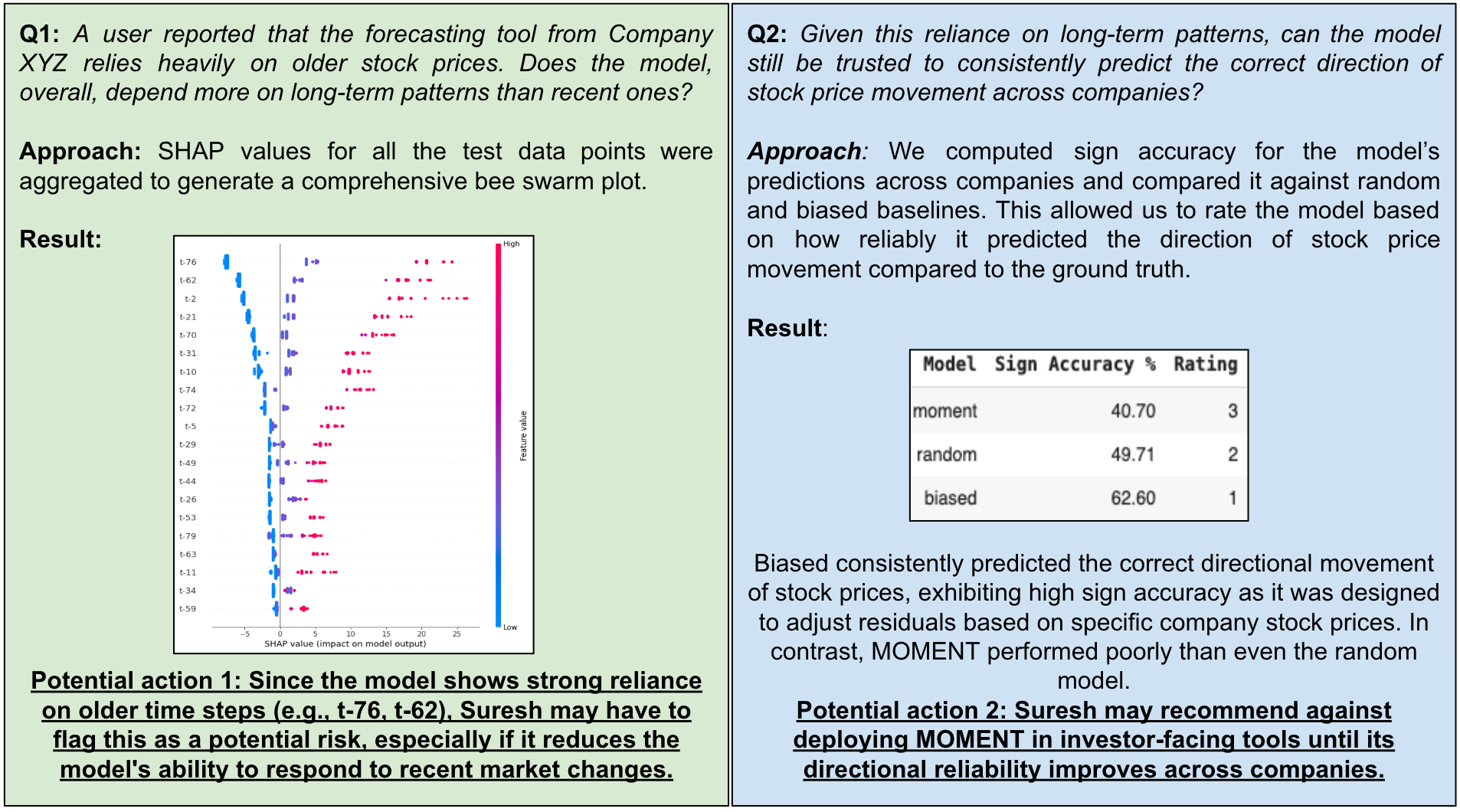}
    \caption{\textbf{Stakeholder:} Suresh (Regulatory official; Regulatory body). \textbf{Scenario-4:} Suresh uses a stock forecasting tool developed by a third-party company to guide her investment decisions. He notices that the tool performs poorly on certain days and suspects it may be less reliable for some companies. To decide whether to continue using it, he wants to understand how robust the tool is.}
    \label{fig:ts-e2}
\end{figure}
\vspace{-1em}

\begin{figure*}
    \centering
    \includegraphics[width=0.75\textwidth, height=6.5cm]{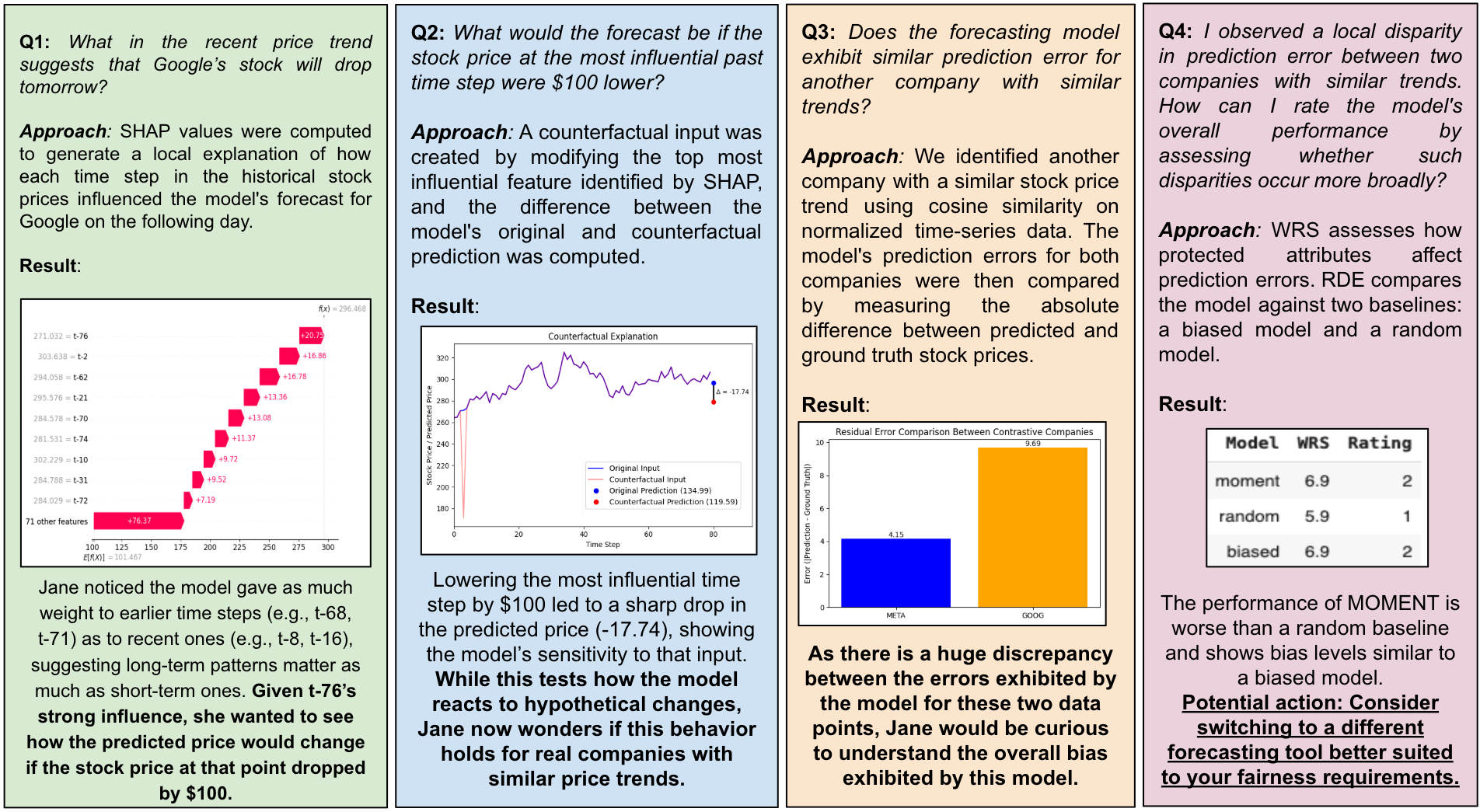}
    \caption{\textbf{Stakeholder:} Jane (Retail Investor; Individual). \textbf{Scenario-5:} Jane uses a stock forecasting tool developed by a third-party company to guide her investment decisions. She notices that the tool performs poorly on certain days and suspects it may be less reliable for some companies. To decide whether to continue using it, she wants to understand how robust the tool is.}
    \label{fig:ts-e1}
\end{figure*}

\begin{figure*}
    \centering
    \includegraphics[width=0.75\textwidth, height=7cm]{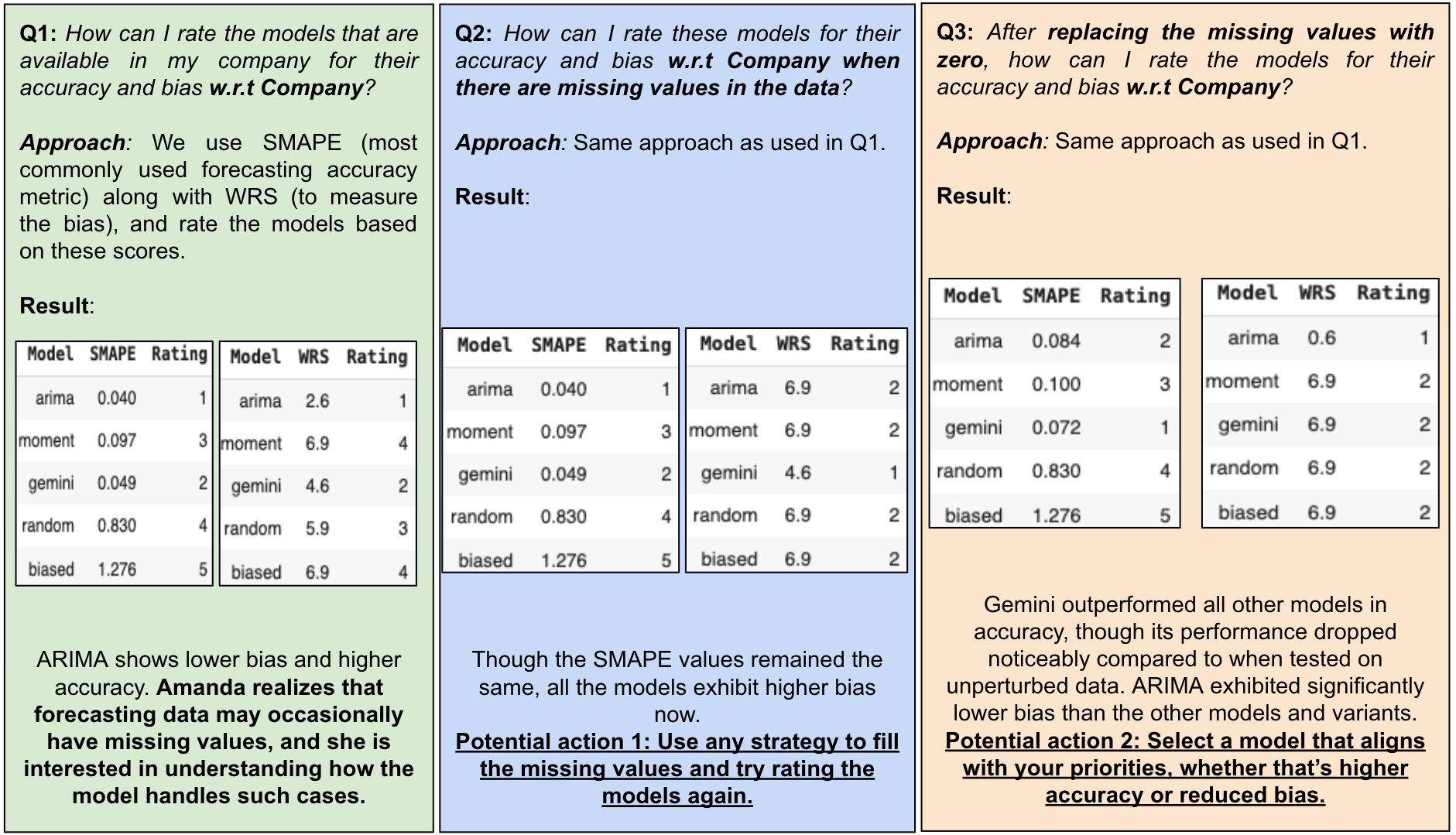}
    \caption{\textbf{Stakeholder:} Amanda (Data scientist; Operational organization). \textbf{Scenario-6:} Amanda is evaluating several time-series forecasting models for deployment. She wants to compare their accuracy and bias to choose one that best fits her team's priorities.}
    \label{fig:ts-e3}
\end{figure*}

\subsection{Case Study-2: Time-Series Forecasting}
\label{sec:case-study-2}
In the second case study, we focus on stock price forecasting, a task that is high-impact in financial decision-making and widely used in algorithmic trading, portfolio management, and investment analytics.

\noindent \textbf{A. Data.}
We collected daily stock prices from Yahoo! Finance for six companies across different industries: Meta (META) and Google (GOO) in social technology, Pfizer (PFE) and Merck (MRK) in pharmaceuticals, and Wells Fargo (WFC) and Citigroup (C) in financial services, as closing stock prices is the most commonly used data source for time-series forecasting \cite{lin2024stock,lakkaraju2024timeseries}. The dataset covers from March 28, 2023, to April 22, 2024; data up to March 22, 2024, was used to forecast the following month.

%\subsubsection{B. Causal Setup.}
\noindent \textbf{B. Causal Setup.}
To generate RDE for time-series forecasting, we adopt the preprocessing approach from \cite{lakkaraju2024timeseries}, where the difference between the 20 predicted stock prices (based on the previous 80 time steps) and their corresponding ground truth values is used to compute a sequence of residuals. While other metrics like SMAPE or MASE are available, we use residuals because they give the raw prediction error at each point, rather than at an aggregate level. To capture worst-case performance, we take the maximum residual within each prediction window as the outcome of interest. For simplicity, we refer to this as the \textit{residual} throughout the paper. While predicted stock prices may vary across companies, our focus is on whether residuals, i.e., model errors, also vary systematically, which may indicate unreliability. This residual serves as the basis for analyzing how protected attributes or perturbations affect model behavior. Figure \ref{fig:ts-cm} shows the causal diagram used for this analysis. We treat \emph{Perturbation} (e.g., introducing missing values or zero-imputation, as in Figure \ref{fig:ts-e3}) as the treatment, \textit{Company} as the sensitive attribute, and \textit{Residual} as the outcome. This causal setup is used exclusively for explanation questions that require RDE.

% \subsubsection{C. AI Models Considered.} 
\noindent \textbf{C. AI Models Considered.}
Our goal is not to benchmark forecasting performance but to analyze and explain model behavior. So rather than selecting the most competitive models, we chose two representative ones from the models used in \cite{lakkaraju2024timeseries}: \textbf{MOMENT}, an open-source, time-series-specific foundation model, and \textbf{Gemini}, a general-purpose foundation model, both of which are easily accessible. We also use \textbf{ARIMA}, a statistical time-series forecasting model, as an additional baseline.

We present three scenarios in \textit{H-XAI} and apply them to address questions posed by different types of stakeholders: a regulatory official (Figure \ref{fig:ts-e2}), a retail investor (Figure \ref{fig:ts-e1}), and a data scientist within an operational organization (Figure \ref{fig:ts-e3}).

\subsection{When to Use: Mapping Questions to Methods}
A key goal of \textit{H-XAI} is to help users in choosing the right tool for the right question, not only to apply methods indiscriminately. %is not just to throw multiple methods at the problem, but to help users choose the right tool for the right question. 
Table~\ref{tab:method-map} maps stakeholder questions to fitting methods. We summarize what each reveals, their limitations, and \textit{when} they require complementary methods. %provides a structured view of which methods work best for different types of stakeholder questions, what each method reveals, and what its limitations are. We also indicate when a method is likely insufficient on its own and should be paired with another. 
This is where \textit{H-XAI} matters: it lets users pose questions, pick methods, identify problems, and cross-validate them using complementary methods. No single method is sufficient; the strength lies in the combination.
% No single method can do it all, and that’s the point. The strength lies in the combination.

\vspace{-0.8em}
\section{Discussion and Conclusion}

Although our case studies focus on credit risk classification and stock price forecasting, \textit{H-XAI} is task-agnostic and extends beyond these domains. The framework centers on selecting explanation techniques based on stakeholder roles and the questions they bring, which is increasingly important as AI systems are deployed through web-based decision services and online platforms. In classification settings, RDE can estimate the causal impact of applicant attributes on model decisions; in forecasting settings, it can quantify how input perturbations affect a model's stability. Traditional XAI methods complement RDE by providing instance-level explanations where causal analysis may not apply.

As shown in Figure \ref{fig:rde-workflow}, RDE is more than a collection of metrics: it provides a step-by-step reasoning process that involves constructing appropriate baselines, outlining a causal model, and choosing the corresponding estimands. This structure allows \textit{H-XAI} to adapt to a variety of stakeholder needs, whether the goal is to investigate robustness, quantify bias, or interpret
individual decisions. These use cases naturally arise in settings where users interact with AI predictions through online interfaces, financial dashboards, or automated decision portals. We recognize that stakeholders may not always know how to specify a causal model. To handle this, RDE adopts a generalized causal model by default, capturing structural relationships that appear across many tasks. Depending on how a question is framed, the analysis may target confounding bias (e.g., DIE\%) or statistical bias (e.g., WRS), as illustrated in Q1 of Figure \ref{fig:gc-e1} versus Q1 of Figure \ref{fig:ts-e3}. Throughout the framework, interpretability is not treated as an inherent property of a single method but as something that emerges from aligning the method with the stakeholder's specific question and context.

A core insight from this work is that explanation is an iterative process rather than a one-off output. Different questions require different tools, and no individual method is adequate on its own. By combining RDE with counterfactual reasoning and feature-based explanations, we shift the role of explanation from post hoc justification to a form of structured interaction that allows stakeholders to
test hypotheses and improve their understanding of model behavior. While this paper presents a first step toward a holistic approach, several directions remain open for future work, including additional case studies, exploring complementary explanation methods, and conducting user studies with diverse stakeholders who rely on AI-driven decisions in online and operational settings.

% -------------
\clearpage
\bibliographystyle{ACM-Reference-Format}
\bibliography{references}

@inproceedings{suresh2021beyond,
  title={Beyond expertise and roles: A framework to characterize the stakeholders of interpretable machine learning and their needs},
  author={Suresh, Harini and Gomez, Steven R and Nam, Kevin K and Satyanarayan, Arvind},
  booktitle={Proceedings of the 2021 CHI conference on human factors in computing systems},
  pages={1--16},
  year={2021}
}

@article{bhatt2020machine,
  title={Machine learning explainability for external stakeholders},
  author={Bhatt, Umang and Andrus, McKane and Weller, Adrian and Xiang, Alice},
  journal={arXiv preprint arXiv:2007.05408},
  year={2020}
}

@article{hoffman2023explainable,
  title={Explainable ai: roles and stakeholders, desirements and challenges},
  author={Hoffman, Robert R and Mueller, Shane T and Klein, Gary and Jalaeian, Mohammadreza and Tate, Connor},
  journal={Frontiers in Computer Science},
  volume={5},
  pages={1117848},
  year={2023},
  publisher={Frontiers Media SA}
}

@article{mueller2019explanation,
  title={Explanation in human-AI systems: A literature meta-review, synopsis of key ideas and publications, and bibliography for explainable AI},
  author={Mueller, Shane T and Hoffman, Robert R and Clancey, William and Emrey, Abigail and Klein, Gary},
  journal={arXiv preprint arXiv:1902.01876},
  year={2019}
}

@incollection{bundas2023facilitating,
  title={Facilitating stakeholder communication around AI-enabled systems and business processes},
  author={Bundas, Matthew and Nadeau, Chasity and Nguyen, Thanh H and Shantz, Jeannine and Balduccini, Marcello and Griffor, Edward and Son, Tran Cao},
  booktitle={Research Handbook on Artificial Intelligence and Communication},
  pages={268--283},
  year={2023},
  publisher={Edward Elgar Publishing}
}

@inproceedings{deshpande2022responsible,
  title={Responsible ai systems: who are the stakeholders?},
  author={Deshpande, Advait and Sharp, Helen},
  booktitle={Proceedings of the 2022 AAAI/ACM Conference on AI, Ethics, and Society},
  pages={227--236},
  year={2022}
}

@ARTICLE{kausik2024rating,
  author={Lakkaraju, Kausik and Srivastava, Biplav and Valtorta, Marco},
  journal={IEEE Transactions on Technology and Society}, 
  title={Rating Sentiment Analysis Systems for Bias Through a Causal Lens}, 
  year={2024},
  volume={},
  number={},
  pages={1-1},
  doi={10.1109/TTS.2024.3375519}
}

@INPROCEEDINGS {kausik2023the,
    author = {K. Lakkaraju and A. Gupta and B. Srivastava and M. Valtorta and D. Wu},
    booktitle = {2023 5th IEEE International Conference on Trust, Privacy and Security in Intelligent Systems and Applications (TPS-ISA)},
    title = {The Effect of Human v/s Synthetic Test Data and Round-Tripping on Assessment of Sentiment Analysis Systems for Bias},
    year = {2023},
    volume = {},
    issn = {},
    pages = {380-389},
    doi = {10.1109/TPS-ISA58951.2023.00053},
    url = {https://doi.ieeecomputersociety.org/10.1109/TPS-ISA58951.2023.00053},
    publisher = {IEEE Computer Society},
    address = {Los Alamitos, CA, USA},
    month = {nov}
}

@inproceedings{srivastava2020personalized,
	title={Personalized Chatbot Trustworthiness Ratings},
	author={Biplav Srivastava and Francesca Rossi and Sheema Usmani and Mariana Bernagozzi},
	booktitle={IEEE Transactions on Technology and Society.},
	year={2020}
}

@inproceedings{srivastava2020rating,
	title={Rating AI Systems for Bias to Promote Trustable Applications},
	author={Biplav Srivastava and Francesca Rossi},
	booktitle={IBM Journal of Research and Development},
	year={2020}
}

@inproceedings{srivastava2018towards,
	title={Towards Composable Bias Rating of AI Systems},
	author={Biplav Srivastava and Francesca Rossi},
	booktitle={2018 AI Ethics and Society Conference (AIES 2018), New Orleans, Louisiana, USA, Feb 2-3},
	year={2018}
}

@ARTICLE{vega-userstudy-translatorbias,
  author={Bernagozzi, Mariana and Srivastava, Biplav and Rossi, Francesca and Usmani, Sheema},
  journal={IEEE Internet Computing}, 
  title={Gender Bias in Online Language Translators: Visualization, Human Perception, and Bias/Accuracy Tradeoffs}, 
  year={2021},
  volume={25},
  number={5},
  pages={53-63},
  doi={10.1109/MIC.2021.3097604}
}

@inproceedings{vega-tool,
  title={Vega: a virtual environment for exploring gender bias vs. accuracy trade-offs in ai translation services},
  author={Bernagozzi, Mariana and Srivastava, Biplav and Rossi, Francesca and Usmani, Sheema},
  booktitle={Proceedings of the AAAI Conference on Artificial Intelligence},
  volume={35},
  number={18},
  pages={15994--15996},
  year={2021}
}

@inproceedings{tian2023mitigating,
author = {Tian, Xinran and Nunes, Bernardo Pereira and Grant, Katrina and Casanova, Marco Antonio},
title = {Mitigating Bias in GLAM Search Engines: A Simple Rating-Based Approach and Reflection},
year = {2023},
isbn = {9798400702327},
publisher = {Association for Computing Machinery},
address = {New York, NY, USA},
url = {https://doi.org/10.1145/3603163.3609043},
doi = {10.1145/3603163.3609043},
booktitle = {Proceedings of the 34th ACM Conference on Hypertext and Social Media},
articleno = {25},
numpages = {5},
location = {Rome, Italy},
series = {HT '23}
}

@article{carey2022causal,
  title={The causal fairness field guide: Perspectives from social and formal sciences},
  author={Carey, Alycia N and Wu, Xintao},
  journal={Frontiers in Big Data},
  volume={5},
  year={2022},
  publisher={Frontiers Media SA}
}

@inproceedings{kausik2022why,
    author = {Lakkaraju, Kausik},
    title = {Why is my System Biased?: Rating of AI Systems through a Causal Lens},
    year = {2022},
    isbn = {9781450392471},
    publisher = {Association for Computing Machinery},
    address = {New York, NY, USA},
    url = {https://doi.org/10.1145/3514094.3539556},
    doi = {10.1145/3514094.3539556},
    booktitle = {Proceedings of the 2022 AAAI/ACM Conference on AI, Ethics, and Society},
    pages = {902},
    numpages = {1},
    location = {Oxford, United Kingdom},
    series = {AIES '22}
}

@InProceedings{srivastava2023advances,
  author = 	"Biplav Srivastava and Kausik Lakkaraju and Mariana Bernagozzi and Marco Valtorta",
  title = 	"Advances in Automatically Rating the Trustworthiness of Text Processing Services",
  booktitle = 	"AI Ethics 4, 5–13. https://doi.org/10.1007/s43681-023-00391-5. Preprint on Arxiv at: https://arxiv.org/abs/2302.09079",
  year = 	"2024",
}

@article{lakkaraju2025creating,
  title={On Creating a Causally Grounded Usable Rating Method for Assessing the Robustness of Foundation Models Supporting Time Series},
  author={Lakkaraju, Kausik and Kaur, Rachneet and Zehtabi, Parisa and Patra, Sunandita and Valluru, Siva Likitha and Zeng, Zhen and Srivastava, Biplav and Valtorta, Marco},
  journal={arXiv preprint arXiv:2502.12226},
  year={2025}
}

@article{lakkaraju2024timeseries,
  title={Rating Multi-Modal Time-Series Forecasting Models (MM-TSFM) for Robustness Through a Causal Lens},
  author={Lakkaraju, Kausik and Kaur, Rachneet and Zeng, Zhen and Zehtabi, Parisa and Patra, Sunandita and Srivastava, Biplav and Valtorta, Marco},
  journal={arXiv preprint arXiv:2406.12908},
  year={2024}
}

@misc{miller2023explainable,
      title={Explainable AI is Dead, Long Live Explainable AI! Hypothesis-driven decision support}, 
      author={Tim Miller},
      year={2023},
      eprint={2302.12389},
      archivePrefix={arXiv},
      primaryClass={cs.AI}
}

@article{hoffman2023increasing,
  title={Increasing the value of XAI for users: A psychological perspective},
  author={Hoffman, Robert R and Miller, Timothy and Klein, Gary and Mueller, Shane T and Clancey, William J},
  journal={KI-K{\"u}nstliche Intelligenz},
  volume={37},
  number={2},
  pages={237--247},
  year={2023},
  publisher={Springer}
}

@article{arrieta2020explainable,
  title={Explainable Artificial Intelligence (XAI): Concepts, taxonomies, opportunities and challenges toward responsible AI},
  author={Arrieta, Alejandro Barredo and D{\'\i}az-Rodr{\'\i}guez, Natalia and Del Ser, Javier and Bennetot, Adrien and Tabik, Siham and Barbado, Alberto and Garc{\'\i}a, Salvador and Gil-L{\'o}pez, Sergio and Molina, Daniel and Benjamins, Richard and others},
  journal={Information fusion},
  volume={58},
  pages={82--115},
  year={2020},
  publisher={Elsevier}
}

@article{arya2019one,
  title={One explanation does not fit all: A toolkit and taxonomy of ai explainability techniques},
  author={Arya, Vijay and Bellamy, Rachel KE and Chen, Pin-Yu and Dhurandhar, Amit and Hind, Michael and Hoffman, Samuel C and Houde, Stephanie and Liao, Q Vera and Luss, Ronny and Mojsilovi{\'c}, Aleksandra and others},
  journal={arXiv preprint arXiv:1909.03012},
  year={2019}
}

@inproceedings{abdul2018trends,
  title={Trends and trajectories for explainable, accountable and intelligible systems: An hci research agenda},
  author={Abdul, Ashraf and Vermeulen, Jo and Wang, Danding and Lim, Brian Y and Kankanhalli, Mohan},
  booktitle={Proceedings of the 2018 CHI conference on human factors in computing systems},
  pages={1--18},
  year={2018}
}

@inproceedings{chromik2021human,
  title={Human-XAI interaction: a review and design principles for explanation user interfaces},
  author={Chromik, Michael and Butz, Andreas},
  booktitle={Human-Computer Interaction--INTERACT 2021: 18th IFIP TC 13 International Conference, Bari, Italy, August 30--September 3, 2021, Proceedings, Part II 18},
  pages={619--640},
  year={2021},
  organization={Springer}
}

@article{gong2024causal,
  title={Causal discovery from temporal data: An overview and new perspectives},
  author={Gong, Chang and Zhang, Chuzhe and Yao, Di and Bi, Jingping and Li, Wenbin and Xu, YongJun},
  journal={ACM Computing Surveys},
  volume={57},
  number={4},
  pages={1--38},
  year={2024},
  publisher={ACM New York, NY}
}

@article{varando2024pairwise,
  title={Pairwise causal discovery with support measure machines},
  author={Varando, Gherardo and Catsis, Salvador and Diaz, Emiliano and Camps-Valls, Gustau},
  journal={Applied Soft Computing},
  volume={150},
  pages={111030},
  year={2024},
  publisher={Elsevier}
}

@inproceedings{zemel2013learning,
  title={Learning fair representations},
  author={Zemel, Rich and Wu, Yu and Swersky, Kevin and Pitassi, Toni and Dwork, Cynthia},
  booktitle={International conference on machine learning},
  pages={325--333},
  year={2013},
  organization={PMLR}
}

@article{frank2010uci,
  title={UCI machine learning repository},
  author={Frank, Andrew},
  journal={http://archive. ics. uci. edu/ml},
  year={2010}
}

@article{demajo2020explainable,
  title={Explainable ai for interpretable credit scoring},
  author={Demajo, Lara Marie and Vella, Vince and Dingli, Alexiei},
  journal={arXiv preprint arXiv:2012.03749},
  year={2020}
}

@book{klosok2020towards,
  title={Towards better understanding of complex machine learning models using explainable artificial intelligence (XAI): Case of credit scoring modelling},
  author={K{\l}osok, Marta and Chlebus, Marcin and others},
  year={2020},
  publisher={University of Warsaw, Faculty of Economic Sciences Warsaw}
}

@article{melsom2022explainable,
  title={Explainable artificial intelligence for credit scoring in banking},
  author={Melsom, Borger and Venner{\o}d, Christian Bakke and de Lange, Petter Eilif},
  journal={Journal of Risk},
  year={2022}
}

@inbook{simpson2022judea,
author = {Pearl, Judea},
title = {Comment: Understanding Simpson’s Paradox},
year = {2022},
isbn = {9781450395861},
publisher = {Association for Computing Machinery},
address = {New York, NY, USA},
edition = {1},
url = {https://doi.org/10.1145/3501714.3501738},
booktitle = {Probabilistic and Causal Inference: The Works of Judea Pearl},
pages = {399–412},
numpages = {14}
}

@article{liao2021question,
  title={Question-driven design process for explainable AI user experiences},
  author={Liao, Q Vera and Pribi{\'c}, Milena and Han, Jaesik and Miller, Sarah and Sow, Daby},
  journal={arXiv preprint arXiv:2104.03483},
  year={2021}
}

@article{atrey2025demystifying,
  title={Demystifying the Accuracy-Interpretability Trade-Off: A Case Study of Inferring Ratings from Reviews},
  author={Atrey, Pranjal and Brundage, Michael P and Wu, Min and Dutta, Sanghamitra},
  journal={arXiv preprint arXiv:2503.07914},
  year={2025}
}

@article{addy2024ai,
  title={AI in credit scoring: A comprehensive review of models and predictive analytics},
  author={Addy, Wilhelmina Afua and Ajayi-Nifise, Adeola Olusola and Bello, Binaebi Gloria and Tula, Sunday Tubokirifuruar and Odeyemi, Olubusola and Falaiye, Titilola},
  journal={Global Journal of Engineering and Technology Advances},
  volume={18},
  number={02},
  pages={118--129},
  year={2024}
}

@article{martins2023explainable,
  title={Explainable artificial intelligence (XAI): A systematic literature review on taxonomies and applications in finance},
  author={Martins, Tiago and De Almeida, Ana Maria and Cardoso, Elsa and Nunes, Lu{\'\i}s},
  journal={IEEE Access},
  volume={12},
  pages={618--629},
  year={2023},
  publisher={IEEE}
}

@article{lin2024stock,
  title={Stock market prediction using artificial intelligence: A systematic review of systematic reviews},
  author={Lin, Chin Yang and Marques, Joao Alexandre Lobo},
  journal={Social Sciences \& Humanities Open},
  volume={9},
  pages={100864},
  year={2024},
  publisher={Elsevier}
}

@article{raykar2023tsshap,
  title={TsSHAP: Robust model agnostic feature-based explainability for time series forecasting},
  author={Raykar, Vikas C and Jati, Arindam and Mukherjee, Sumanta and Aggarwal, Nupur and Sarpatwar, Kanthi and Ganapavarapu, Giridhar and Vaculin, Roman},
  journal={arXiv preprint arXiv:2303.12316},
  year={2023}
}

@article{wachter2017counterfactual,
  title={Counterfactual explanations without opening the black box: Automated decisions and the GDPR},
  author={Wachter, Sandra and Mittelstadt, Brent and Russell, Chris},
  journal={Harv. JL \& Tech.},
  volume={31},
  pages={841},
  year={2017},
  publisher={HeinOnline}
}

@inproceedings{verma2018fairness,
  title={Fairness definitions explained},
  author={Verma, Sahil and Rubin, Julia},
  booktitle={Proceedings of the international workshop on software fairness},
  pages={1--7},
  year={2018}
}
% \clearpage
% \input{sections/checklist}
\clearpage
\appendix
\newpage
\setcounter{section}{0}
\renewcommand{\thesection}{\Alph{section}}
\setcounter{figure}{0}
\renewcommand{\thefigure}{\thesection.\arabic{figure}}
\setcounter{table}{0}
\renewcommand{\thetable}{\thesection.\arabic{table}}

\section*{Appendix}

\section{RDE Metrics}
\label{app:rde-metrics}

Here is how the rating metrics are calculated.
\begin{itemize}
    \item \textbf{Algorithm 1} computes the Weighted Rejection Score (WRS) to measure statistical bias by analyzing how sensitive the model's outcomes are to variations in sensitive attributes.

    \item \textbf{Algorithm 2} calculates Deconfounding Impact Estimation (DIE \%) using PSM to assess confounding bias by measuring the effect of the confounder on model outcomes before and after deconfounding.
    
    \item \textbf{Algorithm 3} calculates ATE by evaluating the difference in the model’s outcomes for treated data (perturbation applied) and control data (no perturbation) to evaluate the impact of the treatment on the outcome. The ATE is computed after applying PSM or G-computation to mitigate the confounding effects, hence the use of the do(.) operator in the ATE formulation.
\end{itemize}

\begin{algorithm}
\small
\caption{\textit{WeightedRejectionScore}}
\label{alg:wrs}
\textbf{Purpose:} Calculate the weighted sum of null-hypothesis rejections for dataset $d_j$ and model $s$ using confidence intervals $CI_k$ and weights $w_k$.

\textbf{Input:} \\
$d$: Dataset for a specific perturbation; \\
$CI$: Confidence intervals (e.g., 95\%, 70\%, 60\%); \\
$s$: AI model from the set of test models $S$; \\
$W$: Weights for different CIs (e.g., 1, 0.8, 0.6) \\
\textbf{Output:} $\psi$: Weighted rejection score

\begin{algorithmic}[1]
\small
\State $\psi \gets 0$
\For{each $(ci_i, w_i)$ in $(CI, W)$}
    \For{each pair $(z_a, z_b)$ in classes of $Z$}
        \State $(t, pval, dof) \gets \text{T-Test}(z_a, z_b)$
        \State $t_{crit} \gets \text{LookUp}(ci_i, dof)$
        \If{$t > t_{crit}$}
            \State $\psi \gets \psi + w_i$
        \EndIf
    \EndFor
\EndFor
\State \Return $\psi$
\end{algorithmic}
\end{algorithm}

\begin{algorithm}
\small
\caption{\textit{ComputeDIEScore}}
\label{alg:pie}
\textbf{Purpose:} Calculate the Deconfounding Impact Estimation (DIE) using Propensity Score Matching (PSM).

\textbf{Input:} \\
$s$: AI model, \\
$d$: Dataset corresponding to a perturbation, \\
$p$: A perturbation other than $p_0$ (control or no perturbation)

\textbf{Output:} $\psi$: DIE score

\begin{algorithmic}[1]
\State $ATE_o \gets E(O \mid P = p) - E(O \mid P = p_0)$ \Comment{Observed APE}
\State $ATE_m \gets E(O \mid do(P = p)) - E(O \mid do(P = p_0))$ \Comment{Deconfounded APE using PSM}
\State $\psi \gets (ATE_m - ATE_o) \times 100$ \Comment{Compute DIE score}
\State \Return $\psi$
\end{algorithmic}
\end{algorithm}

\begin{algorithm}
\small
\caption{\textit{ComputeATEScore}}
\label{alg:ate}
\textbf{Purpose:} Calculate the Average Treatment Effect (ATE).

\textbf{Input:} \\
$s$: AI model, \\
$d$: Dataset, \\
$p$: Treatment perturbation, \\
$p_0$: Control (unperturbed baseline)

\textbf{Output:} $\psi$: ATE score

\begin{algorithmic}[1]
\State $\psi \gets E(O \mid do(P = p)) - E(O \mid do(P = p_0))$ \Comment{Compute ATE}
\State \Return $\psi$
\end{algorithmic}
\end{algorithm}

% \subsection{Traditional XAI Approaches Employed}

% \section{Additional Scenarios from Case Studies}
% \label{app:case-studies}

\begin{table}[h]
\small
\centering
\begin{tabular}{|l|c|c|}
\hline
\textbf{Model} & \textbf{SMAPE (\%)} & \textbf{MASE} \\
\hline
Surrogate & 7.88 & 4.77 \\
MOMENT & 8.06 & 5.45 \\
\hline
\end{tabular}
\caption{MASE and SMAPE scores on a 10\% evaluation subset (60 samples) for the MOMENT model and its surrogate. The small differences indicate that the surrogate closely approximates MOMENT's predictions for explanation purposes.}
\label{tab:mase-smape-surrogate}
\end{table}

\arrayrulecolor{black}

\begin{table*}[t]
    \small
    \centering
    \begin{tabular}{|p{0.15\linewidth}|p{0.47\linewidth}|p{0.3\linewidth}|}
        \hline
        \textbf{Scenario} & \textbf{Question in Paper} & \textbf{Mapped Question from XAI Question Bank} \\
        \hline
        \multicolumn{3}{|c|}{\textbf{Case Study 1: Binary Classification (Credit Risk)}} \\
        \hline
        \rowcolor{green!5}
        Individual (Fig. \ref{fig:gc-e1}) & Q1: On my data instance, I have observed the AI model used by the bank to be biased... How can I rate this model for expected behavior? & \textit{Performance:} How reliable (fairness) are the predictions? \\
        \hline
        Individual (Fig. \ref{fig:gc-e1}) & Q2: How did each factor... contribute to my loan rejection? & \textit{Why:} What feature(s) of this instance determine the system’s prediction
        of it\\
        \hline
        Individual (Fig. \ref{fig:gc-e1}) & Q3: What is the smallest change I could make to my loan duration... to get my loan approved? & \textit{How to be that:} What is the minimum change required for this instance to get a different prediction?  \\
        \hline
        Individual (Fig. \ref{fig:gc-e1}) & Q4: What is the smallest change I could make to my checking account balance to get my loan approved? & \textit{How to be that:} What is the minimum change required for this instance to get a different prediction?  \\
        \hline
        \rowcolor{green!5}
        Regulatory (Fig. \ref{fig:gc-e2}) & Q1: An applicant reported that the model... is biased w.r.t personal status and gender. How can I rate this model for expected behavior? & \textit{Performance:} How reliable (fairness) are the predictions?\\
        \hline
        \rowcolor{green!5}
        Regulatory (Fig. \ref{fig:gc-e2}) & Q2: Overall, how does the predicted risk change with the applicant's personal status and gender? & \textit{How:} How does the system make predictions? \\
        \hline
        \rowcolor{green!5}
        Regulatory (Fig. \ref{fig:gc-e2}) & Q3: How does the model treat female applicants compared to male applicants with varying personal/marital status differently? & \textit{How:} How does the system make predictions? (Specifically, its behavior across subgroups) \\
        \hline
        Data Scientist (Fig. \ref{fig:gc-e3}) & Q1: How can I rate the models that are available in my company for accuracy? & \textit{Performance:} How accurate are the predictions?  \\
        \hline
        \rowcolor{green!5}
        Data Scientist (Fig. \ref{fig:gc-e3}) & Q2: How can I rate the models for their bias w.r.t personal status, gender, and age? & \textit{Performance:} How reliable (fairness) are the predictions? \\
        \hline
        Data Scientist (Fig. \ref{fig:gc-e3}) & Q3: On average, how does each feature influence the predicted credit risk in the logistic regression model? & \textit{How:} How does the system make predictions? \\
        \hline
        \multicolumn{3}{|c|}{\textbf{Case Study 2: Time-Series Forecasting}} \\
        \hline
        Retail Investor (Fig. \ref{fig:ts-e1}) & Q1: What in the recent price trend suggests that Google's stock will drop tomorrow? & \textit{Why:} What features of this instance determine the system's prediction of it?\\
        \hline
        Retail Investor (Fig. \ref{fig:ts-e1}) & Q2: What would the forecast be if the stock price at the most influential past time step were \$100 lower? & \textit{What If:} What would the system predict if this instance changes to...? \\
        \hline
        \rowcolor{green!5}
        Retail Investor (Fig. \ref{fig:ts-e1}) & Q3: Does the forecasting model exhibit similar prediction error for another company with similar trends? & \textit{Why: }Why are [instance A and B] given the same prediction? \\
        \hline
        \rowcolor{green!5}
        Retail Investor (Fig. \ref{fig:ts-e1}) & Q4: I observed a local disparity in prediction error between two companies... How can I rate the model's overall performance by assessing whether such disparities occur more broadly? & \textit{Performance:} How reliable (fairness) are the predictions? \\
        \hline
        Regulatory (Fig. \ref{fig:ts-e2}) & Q1: A user reported that the forecasting tool... relies heavily on older stock prices. Does the model overall, depend more on long-term patterns than recent ones? & \textit{How:} How does it weigh different features? \\
        \hline
        \rowcolor{green!5}
        Regulatory (Fig. \ref{fig:ts-e2}) & Q2: Given this reliance on long-term patterns, can the model still be trusted to consistently predict the correct direction of stock price movement across companies? & \textit{Performance:} What are the limitations of the system? (specifically, its reliability)  \\
        \hline
        \rowcolor{green!5}
        Data Scientist (Fig. \ref{fig:ts-e3}) & Q1: How can I rate the models that are available... for their accuracy and bias w.r.t Company? & \textit{Performance:} How accurate / reliable (fairness) are the predictions?\\
        \hline
        \rowcolor{green!5}
        Data Scientist (Fig. \ref{fig:ts-e3}) & Q2: How can I rate these models for their accuracy and bias w.r.t Company when there are missing values in the data? & \textit{Performance:} How reliable (fairness and sensitivity to input perturbations (missing values)) are the predictions? \\
        \hline
        \rowcolor{green!5}
        Data Scientist (Fig. \ref{fig:ts-e3}) & Q3: After replacing the missing values with zero, how can I rate the models for their accuracy and bias w.r.t Company? & \textit{Performance:} How reliable (fairness and sensitivity to input perturbations (zeros)) are the predictions?  \\
        \hline
    \end{tabular}
    \caption{Mapping of all scenario questions to the XAI Question Bank from \cite{liao2021question}. The highlighted questions in the table are specifically focused on model fairness and sensitivity to perturbations. While the question bank lacks explicit questions for these topics, we have mapped our questions to the closest available ones, highlighting a need for the original taxonomy to be extended to include these topics.}
    \label{tab:question_mapping}
\end{table*}

Figures \ref{fig:gc-e2},\ref{fig:gc-e3} show additional scenarios from case studies illustrated in Section \ref{sec:cases}. Table \ref{tab:question_mapping} shows how the questions from the paper's case studies align with an existing XAI taxonomy \cite{liao2021question}. The table highlights specific questions on model fairness and sensitivity to perturbations, which lack explicit counterparts in the question bank.

\end{document}